\documentclass[10pt,twocolumn,letterpaper]{article}

\usepackage{iccv}
\definecolor{iccvblue}{rgb}{0.21,0.49,0.74}

\usepackage[pagebackref,breaklinks,colorlinks,allcolors=iccvblue]{hyperref}
\usepackage{multirow}
\usepackage{colortbl}
\usepackage{diagbox}
\usepackage{tabularx}
\usepackage{appendix}
\usepackage{xr}     

\def\paperID{7351} % *** Enter the Paper ID here
\def\confName{ICCV}
\def\confYear{2025}

\title{OVG-HQ: Online Video Grounding with Hybrid-modal Queries}

\author{
Runhao Zeng$^{1\text{*}}$, Jiaqi Mao$^{1}$\thanks{Equal contribution}, Minghao Lai$^{1}$, Minh Hieu Phan$^{2}$,\\ Yanjie Dong$^{1}$,
Wei Wang$^{1}$, Qi Chen$^{2\dagger}$, Xiping Hu$^{1\dagger}$ \\
$^1$Artificial Intelligence Research Institute, Shenzhen MSU-BIT University, $^2$University of Adelaide\\
{\tt\small zengrh@smbu.edu.cn, maojiaqi2324@gmail.com, huxp@smbu.edu.cn}}

\begin{document}
\maketitle

\begingroup
\renewcommand\thefootnote{$\dagger$}
\footnotetext{Corresponding authors}
\endgroup

\begin{abstract}
Video grounding (VG) task focuses on locating specific moments in a video based on a query, usually in text form. However, traditional VG struggles with some scenarios like streaming video or queries using visual cues. To fill this gap, we present a new task named Online Video Grounding with Hybrid-modal Queries (OVG-HQ), which enables online segment localization using text, images, video segments, and their combinations. This task poses two new challenges: limited context in online settings and modality imbalance during training, where dominant modalities overshadow weaker ones. To address these, we propose OVG-HQ-Unify, a unified framework featuring a Parametric Memory Block (PMB) that retain previously learned knowledge to enhance current decision and a cross-modal distillation strategy that guides the learning of non-dominant modalities. This design enables a single model to effectively handle hybrid-modal queries. Due to the lack of suitable datasets, we construct QVHighlights-Unify, an expanded dataset with multi-modal queries. Besides, since offline metrics overlook prediction timeliness, we adapt them to the online setting, introducing oR@$n$, IoU=$m$, and online mean Average Precision (omAP) to evaluate both accuracy and efficiency. Experiments show that our OVG-HQ-Unify outperforms existing models, offering a robust solution for online, hybrid-modal video grounding. Source code and datasets are available at \url{https://github.com/maojiaqi2324/OVG-HQ}.
\end{abstract}    
\section{Introduction}
\label{sec:intro}

\begin{figure}[t]
    \includegraphics[width=\columnwidth]{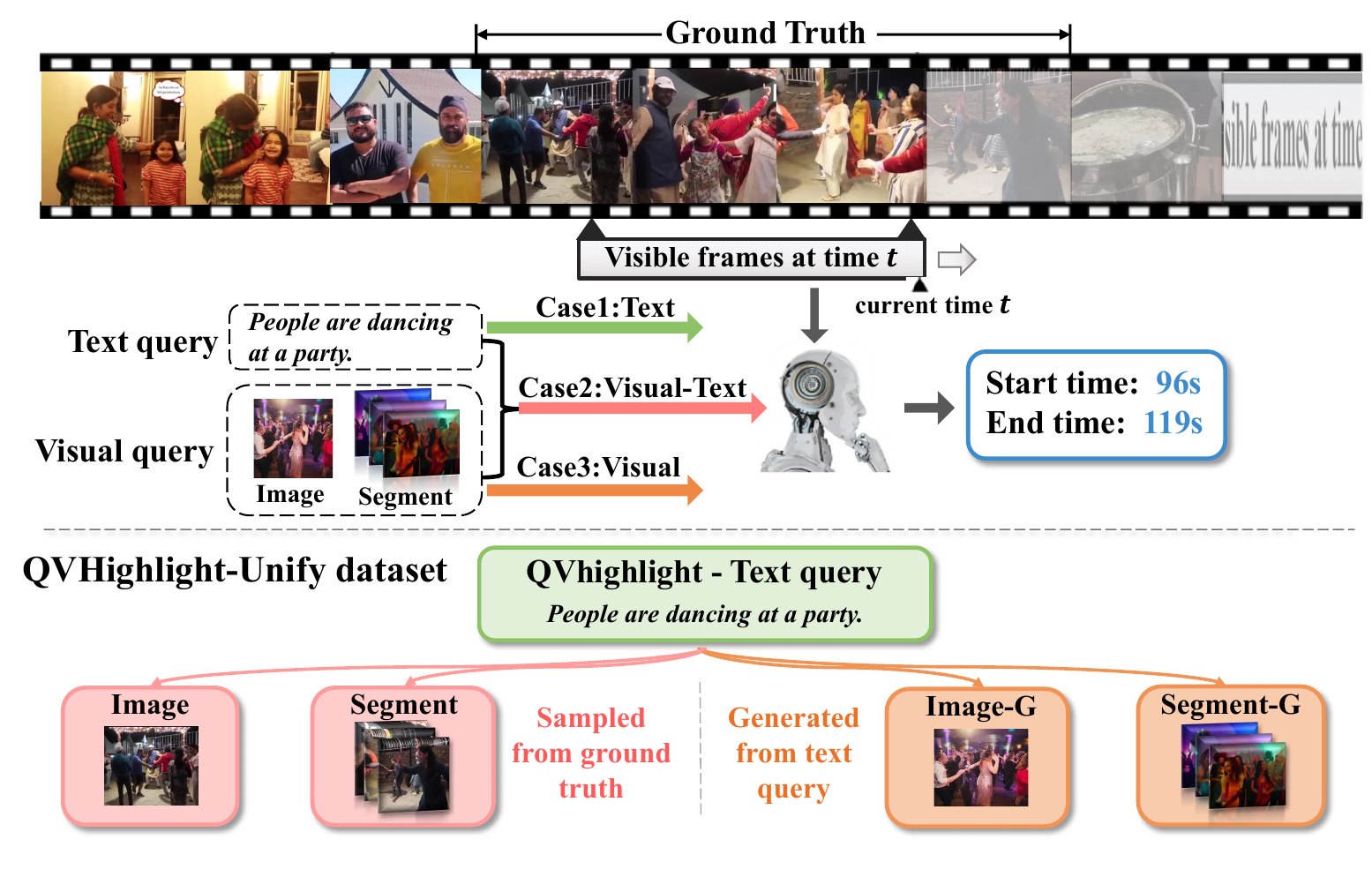}
    \vspace{-20pt}
    \caption{Illustration of our proposed online video grounding with hybrid-modal queries task, with two distinguishing characteristics: online video input and various query configurations. Beyond text query, it accepts visual queries (images, video segments) and their combination with text. We also construct a new QVHighlights-Unify dataset by augmenting QVHighlights dataset with images and video segments and complementary image-text pairs.
    }
    \label{fig:task}
    \vspace{-10pt}
\end{figure}

Video grounding~\cite{gao2017tall,zheng2025training,zeng2025unimd} is a crucial research task that identifies the start and end times of target segments in untrimmed videos based on a given query.
However, the current setting suffers from two critical limitations in real-world applications.
\textbf{First}, it considers an offline setting, imposing strict requirements on complete video accessibility, which is insufficient for immediate detection in streaming media. For example, in surveillance, we need to continuously analyze live feeds and instantly ground queries, such as \textit{``group of people gathering near the front door''}, so that security teams can respond immediately, rather than waiting to process a lengthy offline recording.
\textbf{Second}, the current video grounding task predominantly relies on natural language queries, limiting its application to multi-modal scenarios.
As an example, a text-only system might demand a detailed description such as \textit{``a group of individuals congregating near the front door, frequently looking around, and making brief contact before dispersing in different directions''}. In contrast, with multi-modal queries, security staff could directly upload a past surveillance clip illustrating similar suspicious behavior.
With this consideration, we introduce an extended task called Online Video Grounding with Hybrid-modal Queries (OVG-HQ).

Unlike the conventional offline video grounding task that only considers text queries as inputs, our OVG-HQ task accommodates multiple query modalities (\eg, text, image, video) in online video streaming, as shown in Figure~\ref{fig:task}. This setup requires the model to dynamically process and integrate information from diverse sources, adapting to evolving queries throughout the video. OVG-HQ emphasizes online inference and cross-modal interactions, challenging the model to ground relevant moments accurately across varying contexts and query types.

The new task poses new challenges.
\textbf{First}, the video content can vary significantly over time, with changing scenes, lighting, and objects. Models must adapt to this variability in a streaming video, accounting for concept drift without losing prior learned knowledge.
\textbf{Second}, as noted by~\cite{zhou2023intra}, modality imbalance poses a significant challenge in hybrid-modal queries, as different modalities (such as text, image, or video) contribute unevenly to the task. Stronger modalities with more informative signals often dominate, overshadowing weaker ones. This imbalance causes the model to rely heavily on stronger modalities, leading to underutilization of the weaker ones, which reduces their contribution and ultimately impacts the model's overall accuracy in integrating diverse information. Consequently, it becomes difficult to use a single unified model to handle all modalities effectively.
To tackle the challenging OVG-HQ task, we propose a unified yet flexible model called OVG-HQ-Unify, which supports hybrid-modal query inputs (\ie, both uni- and multi-modal queries) and enables online localization of moments. It mainly has two parts. First, since each streaming video can be regarded as a sequence, to retain previously learned knowledge, we embed a \textbf{Parametric Memory Block} (PMB) instantiated with Test-Time Training layer (TTT)~\cite{sun2024learning} that uses the network’s parameters as dynamic memory for sequence modeling. With a self-supervised reconstruction loss, PMB encodes historical feature and prediction information, allowing the model to ``memorize'' past context for better decisions rather than directly saving historical data. In online video streams, PMB’s ability to update parameters during inference enables continuous improvement and adaptability to new scenarios.
Second, to alleviate the impact of modality imbalance, we design a hybrid distillation strategy that introduces a teacher model to guide the learning of non-dominant modalities, thus enhancing the model’s performance consistency across different query modalities.

As there is no off-the-shelf dataset suitable for the OVG-HQ task, we construct a new dataset called QVHighlights-Unify, which expands the QVHighlights dataset~\cite{lei2021detecting} by adding image and segment queries\footnote{We first expand the QVHighlights for its well-annotated moment retrieval data, enabling systematic evaluation of hybrid-modal queries. In the future, we will collect more complex datasets (\eg, surveillance videos) to further validate and enhance our model in more practical scenarios.}.
This expansion enables the model to handle not only text queries but also visual modality inputs, validating its adaptability and consistency across various query types.
Besides, as the offline metrics fail to capture the timeliness of predictions, we adapt them to the online setting called oR@$n$, IoU=$m$ and online mean Average Precision (omAP) to evaluate both accuracy and efficiency.
Experiments on QVHighlights-Unify, ANet-Captions, TACoS, MAD datasets show that our OVG-HQ-Unify framework achieves superior performance compared to existing methods, particularly in handling hybrid-modal queries.
Our main contributions are as follows:
\begin{itemize}
    \item We introduce a new task, Online Video Grounding with Hybrid-modal Queries (OVG-HQ), enabling multi-modal queries and requiring online segment localization in video streams, which is suited for practical applications.
    \item We propose a unified framework, called OVG-HQ-Unify, supporting hybrid-modal queries as inputs and enabling online localization of video clips. In detail, we introduce a Parameter Memory Block (PMB) to keep previously learned knowledge and a cross-modal distillation strategy to mitigate imbalances during multi-modal training.
    \item We construct a new dataset, QVHighlights-Unify, which includes multiple query modalities. Experiments on 4 datasets show that our OVG-HQ-Unify framework outperforms existing models, demonstrating its superiority in the online setting across various query types.
\end{itemize}

\section{Related Work}
\label{sec:related_work}

\subsection{Video Grounding with Text Query}\label{sec:Video Grounding}
\noindent \textbf{Offline Setting.}
Offline Video Grounding methods~\cite{yang2021deconfounded,wang2022negative,mun2020local,liu2022memory,jang2023knowing,luo2023towards,huang2022video,gao2017tall,croitoru2023moment,zheng2025training,zeng2025unimd,yang2024task,xiao2024bridging,qu2024chatvtg,mu2024snag,fang2024rethinking} involve identifying time intervals within a video that are semantically aligned with a given sentence. Proposal-based methods typically follow a two-stage pipeline: the first stage generates proposals, and the second ranks these proposals based on their relevance to the input query. Early techniques generate proposals using sliding windows~\cite{ge2019mac, zhang2019exploiting, gao2017tall} or predefined temporal anchors~\cite{zhang2019cross, zhang2019man, yuan2019semantic, xu2019multilevel, chen2018temporally}. Later methods~\cite{zhang2020learning, xiao2021boundary, wang2022negative, liu2021context, li2023g2l} explore all possible pairs of start and end points or use 2D temporal maps to process multiple candidates at once. Proposal-free methods~\cite{zhang2020span, ghosh2019excl, yuan2019find, mun2020local} aim to predict the target moment directly without the need for explicit proposals.
They learn the interaction between video and sentence by applying techniques like attention mechanisms~\cite{mun2020local, ghosh2019excl, zhang2020span, yuan2019find, rodriguez2020proposal} and dense regression~\cite{zeng2020dense, lu2019debug, chen2020rethinking} from individual frames. In addition, efforts have been made to integrate temporal sentence grounding with other video understanding tasks into unified frameworks~\cite{yan2023unloc, lin2023univtg}. Recent query-based models~\cite{lei2021detecting, sun2024tr, moon2023query, moon2023correlation, lee2025bam, jang2023knowing, xu2024mh, cao2021pursuit, liu2022umt, li2024momentdiff} have simplified the process by removing the need for handcrafted components. Training-free methods~\cite{xu2024vtg, luo2024zero} have been introduced to address challenges in supervised learning, such as biases from annotations and limited generalization. They avoid relying on annotated data and instead leverage pre-trained models to assess the similarity between video segments and textual queries. Some methods~\cite{luo2024zero} use vision-language models, while others~\cite{xu2024vtg} utilize large language models to compare video frame captions with the query. However, all these methods assume full access to the video in advance, which is not feasible for streaming applications where predictions must be made in streaming videos.

\noindent \textbf{Online Setting.}
Recently,~\cite{gan2023temporal} proposed video grounding in an online setting, which involves retrieving relevant moments given a language query during video streaming. However, this setting overlooks the inherent flexibility of the query itself, as users may require inputs from multiple modalities beyond text, such as image, video segments, or any combination of these modalities.
In this paper, we propose a task that is more aligned with real-world application scenarios, called Online Video Grounding with Hybrid-modal Queries. This task enables online segment localization in video streams using hybrid-modal queries, accommodating various input modalities to better meet user needs.

\subsection{Video Grounding with Multi-modal Query}
Video grounding tasks involve localizing specific events or activities within videos based on a given query. Most methods~\cite{yang2021deconfounded,wang2022negative,mun2020local,liu2022memory}  use natural language as the query. ~\cite{zhang2019localizing} was the first to utilize image queries to localize unseen activities in videos. More recently, \cite{goyal2023minotaur} proposed grounding videos spatio-temporally using images or texts. \cite{zhang2024localizing} attempts to localize events in videos using multimodal semantic queries, but image-text pairs in this dataset are complementary and cannot be used independently as queries, neglecting that users may input different types of queries in practical settings.
In this paper, we unify multiple modalities and various combinations of queries and additionally introduce the concept of video segment queries, which enables segment localization in video streams using queries comprising any combination of modalities—including images, text, and video segments.

\section{Proposed Method}
\label{sec:proposed_method}

\subsection{Problem Definition} 
\noindent \textbf{Offline Video Grounding with Text Query.} This conventional task requires a machine to process an untrimmed video \( V = \{x_i\}_{i=1}^{T} \), where \( x_i \) denotes the \( i \)-th frame, and subsequently identify \( M \) relevant moments \( \mathcal{M} = \{\mathcal{M}_m = (s_m, e_m)\}_{m=1}^{M} \) that correspond to a text query \( \mathcal{Q} \). Each moment \( \mathcal{M}_m \) is defined by its start and end frames \( s_m \) and \( e_m \). However, this offline setting has two primary limitations in practical applications: 1) videos are often streamed, rendering it impractical to wait until all frames have been processed before predicting moments; 2) users may require inputs from multiple modalities beyond text, such as images or video segments.

\noindent \textbf{Online Video Grounding with Hybrid-modal Queries (OVG-HQ).}
In this paper, we propose to study a more practical setting, which aims to understand an input multi-modal query \( \mathcal{Q} \subseteq \{q_t, q_i, q_s\} \)—where \( q_t \), \( q_i \), and \( q_s \) represent text, image, and video segment queries, respectively—and retrieve relevant moments from streaming video.
In this setting, at each timestamp \( t \) (\( 1\leq t\leq T \)), the model only has access to a sliding window of frames\footnote{Accessing all past frames is ideal but impractical for long video streams due to computational and memory constraints. A sliding window offers a balanced trade-off between efficiency and accuracy.} \( V_{t-k+1:t} = \{x_i\}_{i=t-k+1}^{t} \), with \( k \geq 1 \).
Using this partial video segment and multi-modal query \( \mathcal{Q} \), the model should identify events (sometimes more than one) relevant to \( \mathcal{Q} \). 
Importantly, once predictions are made at any timestamp, they cannot be modified or removed in future steps. Current methods rely on Non-Maximum Suppression (NMS) and future frame predictions to adjust past frames, which is impractical in streaming settings.

\begin{figure*}[t]
    \includegraphics[width=\textwidth]{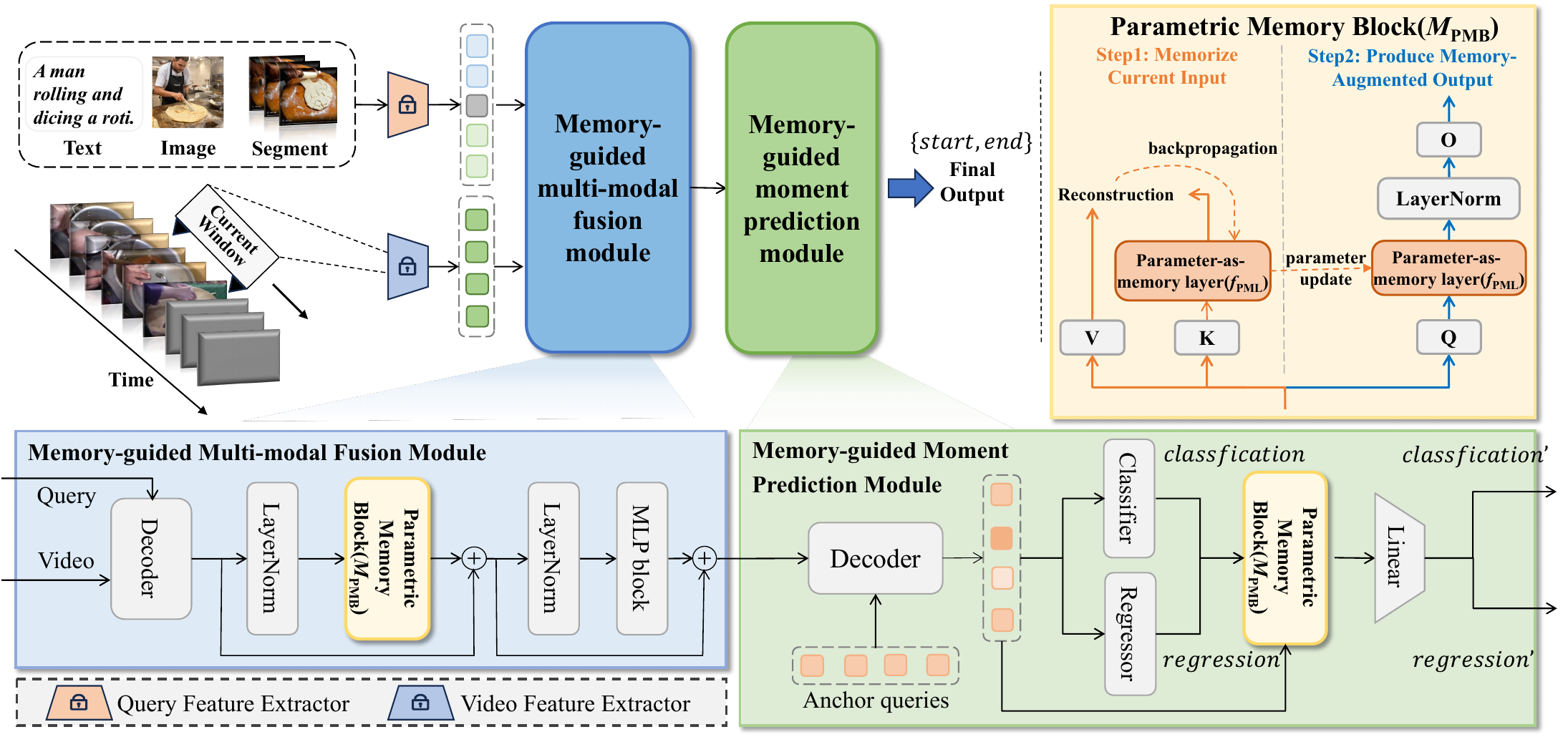}
    \caption{Overview of our OVG-HQ-Unify model. At time \(t\), features extracted from video and query are processed via the memory-guided multi-modal fusion module (Sec.~\ref{Sec:PML_fusion}), where query-aware features are extracted via a transformer decoder and enhanced by parametric memory block \(M_{\text{PMB}}\) (Sec.~\ref{Sec:PMB}). Then, the memory-guided moment prediction module (Sec .~\ref {Sec:PML_prediction}) decodes anchor features, which, along with the current predictions are fed to \(M_{\text{PMB}}\) for moment prediction. In \(M_{\text{PMB}}\), the parameter-as-memory layer (\(f_{\text{PML}}\)) first memorizes current input by updating its parameters via self-supervised reconstruction loss and then predicts based on the historical information.} 
    \label{Fig.2_model}
\end{figure*}

\subsection{General Scheme}

The challenge of online video grounding (OVG) lies in how to efficiently model and utilize historical information to enhance current predictions. To address this, we propose a simple yet effective sequence modeling module, namely \textbf{parametric memory block} (\(M_\text{PMB}\)). Inspired by TTT~\cite{sun2024learning}, our \textbf{parameter-as-memory layer} \(f_{\text{PML}}\) in \(M_\text{PMB}\) compresses sequential information (e.g., input frame sequences) into the neural network parameters. Based on \(M_\text{PMB}\), we design an OVG-HQ-Unify model capable of handling various input configurations, including text, text + image, and text + segment, as shown in Figure \ref{Fig.2_model}.

In the following, we first introduce the design of \(M_\text{PMB}\) in Sec. \ref{Sec:PMB}. We then illustrate how \(M_\text{PMB}\) is employed in multi-modal fusion and prediction in Sec. \ref{Sec:PML_fusion} and \ref{Sec:PML_prediction}, respectively. Lastly, we describe the approach for training a unified model with hybrid-modal queries in Sec. \ref{Sec:Unified_training}.

\subsection{Parametric Memory Block}
\label{Sec:PMB}
To enable memory retention in models, one common approach is to use a memory bank and integrate current inputs with stored memory via self-attention~\cite{wu2022memvit}. However, this introduces extra storage overhead and results in increased computational costs as the amount of historical data grows. In contrast, LSTMs store historical information in a fixed-size hidden state, whose expressive capacity is limited~\cite{sun2024learning}. Unlike the above approaches, we propose a learnable parametric memory block $M_{\textrm{PMB}}$ instantiated with TTT~\cite{sun2024learning} that can compress the historical information within
network parameters, which have much stronger expression power as neural networks have a larger capacity than the hidden states of LSTM. It operates in two steps, as shown in Figure \ref{Fig.2_model}.

\noindent\textbf{Step 1: Memorize Current Input.} 
The core component of $M_{\textrm{PMB}}$ is the \textit{parameter-as-memory layer} \( f_{\textrm{PML}}(\cdot; W^m) \). To compress the current input \( r_t \) into \( W^m \), we employ a reconstruction loss as a form of self-supervision. This approach is akin to how language models utilize reconstruction or masked prediction loss to embed knowledge from training data into the parameters of neural networks through gradient descent. Formally, the reconstruction loss can be defined as follows:
\begin{eqnarray}\label{eq:reconstruction_loss}
    \mathcal{L}_\textrm{{PML}}(r_t;W^m) = \parallel f_\textrm{{PML}}(W_K r_t; W^m) - W_V r_t \parallel^2,
    \label{eqn:gml}
\end{eqnarray}
where $W_K$ and $W_V$ are two learnable projection matrices. We then update $W^m$ by
\begin{eqnarray}\label{eq:update_w}
W^m\leftarrow W^m - \eta_\textrm{{PML}} \cdot \nabla \mathcal{L}_\textrm{{PML}}(r_t;W^m),
\end{eqnarray}
where $\eta_\textrm{{PML}} = \sigma (W_{lr} \cdot r_t)$ is an adpative learning rate following~\cite{sun2024learning}, $W_{lr}$ is a learnable vector and $\sigma$ is the sigmoid function. 
At this point, $W^m$ holds information from both prior and current time step, enabling the network parameters to retain the current representation effectively.

\noindent\textbf{Step 2: Produce Memory-Augmented Output.} 
With the updated memory capturing both current and historical information, we can now augment $r_t$ with memory. The current input \(r_t\) is first processed through a projection layer \(W_Q\), then passed through the updated function \(f_{\textrm{PML}}(\cdot; W^m)\), followed by layer normalization and another projection layer \(W_O\). Mathematically, the process can be defined as
\begin{eqnarray}\label{eq:correct_gradient}
    \hat{r}_t = f_{\textrm{PML}}(r_t;W) = W_O \cdot \textrm{LN} (f_\textrm{{PML}}(W_Q r_t;W^m)),
    \label{eqn:predict}
\end{eqnarray}
where $\textrm{LN}$ denotes a LayerNorm layer. Then, this memory-augmented $\hat{r_t}$ is forwarded to the consequent modules.

\noindent \textbf{Update Rule of Parametric Memory Block.} Let $W^{p}$ be the parameters of $M_{\textrm{PMB}}$, by excluding those of $f_{\textrm{PML}}(\cdot; W^m)$, we denote the remaining parameters as \textbf{\(W^{r} = W^p \setminus W^m\)}.
In other words, all these parameters \(W_Q\), \(W_K\), \(W_V\)and \(W_O\) belong to \(W^{r}\), as illustrated in the upper-right section of Figure~\ref{Fig.2_model}.
\textbf{First}, fix the parameters \(W^r\), forward the current input $r_t$ into $f_\textrm{{PML}}$ and use Eqn. (\ref{eqn:gml}) to update the $f_\textrm{{PML}}$ parameters \(W^m\). \textbf{Second}, with parameters \(W^m\) fixed, use Eqn. (\ref{eqn:predict}) to produce memory-augmented output. \textbf{Third}, update the parameters \( W^r \) by minimizing the loss function derived from the video grounding task.

\subsection{Memory-guided Multi-Modal Fusion}\label{Sec:PML_fusion}

\noindent\textbf{Query Feature Extraction.}

For text and image queries, we use the text and image encoder of CLIP~\cite{radford2021learning} to extract features \( \mathbf{F}_t \) and \( \mathbf{F}_i \), respectively. For segment queries, we use~\cite{radford2021learning} to extract features \( \mathbf{F}_s \) at intervals of \( M \) seconds.

\noindent\textbf{Video Feature Extraction.}
We process video sequences as streaming data through a sliding window mechanism with size \( L \), which dynamically emulates the model's temporal receptive field at each time instant \( t \) by spanning frames within the interval \([t - L, t]\). The window slides forward with a step size of \( M \) seconds, where features of overlapping segments are computed only once and cached for subsequent reuse. Consequently, at each temporal position \( t \), we employ~\cite{radford2021learning} to extract new features from the current video frame. This operational paradigm ultimately yields snippets-level features \( \mathbf{F}_v \in \mathbb{R}^{K \times D_v} \) for each sliding window, where \( K \) denotes the number of video frames extracted within the window.  

\noindent \textbf{Transformer-based Cross-modal Fusion.}
We transform all unimodal features to a unified dimension \( D \) via modality-specific linear layers and use a Transformer decoder with cross-attention to fuse video and query features. Queries may include multiple modalities, so we pad each modality with a specific token \( \mathbf{m}_* \), where \( * \in \{t, i, s\} \). For example, with text and image queries, the decoder input is a combination of multi-modal features \( \mathbf{Q} = [\mathbf{m}_t, \mathbf{F}_t, \mathbf{m}_i, \mathbf{F}_i] \). 
In the decoder, video snippets' features \( \mathbf{F}_v \) in each window serve as queries \( Q_v \), and query features $\mathbf{Q}$ serve as keys \( K_q \) and values \( V_q \). The rest of the decoder follows the standard Transformer architecture, resulting in query-aware video representations \( \mathbf{F}_{qv} \).

\noindent \textbf{Memory-guided Fusion via $f_\text{{PML}}$.}
As the query-aware video feature \(\mathbf{F}_{qv}\) mainly focuses on information within the current window,
to capture long-term video relationships, we further introduce a memory-guided sequence modeling module based on \(f_\text{PML}\) to incorporate historical context. As shown in Figure \ref{Fig.2_model}, this module resembles a Transformer encoder but replaces the self-attention layer with our \(f_\text{PML}\) mechanism.
At each time step \(t\), the feature vector \(\mathbf{F}_{qv}\) is processed by our new module,
and produces an output according to the equation in Eqn.~(\ref{eqn:predict}). This update merges current and historical information, producing a memory-guided feature \(\mathbf{\hat{F}}_{qv}\) for subsequent moment prediction.

\subsection{Memory-guided Moment Prediction}\label{Sec:PML_prediction}

 At time \( t \), our model generates a series of proposals based on predefined anchors, which end at \( t \) with lengths \( L_n = L_q / 2^{n-1} \) for \( n = 1, \dots, N \). For instance, the \( n \)-th anchor is represented as \( A_n = (t - L_n, t) \). We use a Transformer decoder structure, following~\cite{kim2022sliding}, to process the learnable anchor query \( \mathbf{A} \in \mathbb{R}^{N \times D} \) and features \(\mathbf{\hat{F}}_{qv}\) from the Memory-guided Multi-Modal Fusion Module, producing anchor features \( \mathbf{F}_a \in \mathbb{R}^{N \times D} \) (see Figure  \ref{Fig.2_model}). Using \( \mathbf{F}_a \), a classification head predicts \( \{ s_f, s_b \} \) for foreground and background scores, while a regression head predicts \( \{ \Delta l, \Delta o \} \), indicating the target moment length and offset. Thus, the \( n \)-th anchor boundary, \( \hat{A}_n = (s_n, e_n) \), is adjusted by:
\begin{equation}\label{sted}
\begin{aligned}
    s_n &= e_n - L_n \exp(\Delta l_n), \\
    e_n &= t + L_n \Delta o_n.
\end{aligned}
\end{equation}

\noindent \textbf{Memory-guided Prediction Refinement.} 
In the online video grounding setting, predictions made at earlier time steps cannot be adjusted later. Thus, we design the model to refine current predictions using past results. As discussed in Sec.~\ref{Sec:PMB}, \( f_\text{PML} \) can retain historical data, inspiring our \textit{Prediction Refinement Module (PRM)}, shown in Fig.~\ref{Fig.2_model}. First, we concatenate the classification outputs \( \{s_f, s_b\} \) and regression outputs \( \{\Delta l, \Delta o\} \), passing them through a linear layer to create the prediction feature \( \mathbf{F}_p \). This is then combined with anchor features \( \mathbf{F}_a \) to form \( \mathbf{F}_c \), which is processed through \( M_{\text{PML}} \).

Within $f_\text{{PML}}$, two main operations occur: 1) The anchor feature and current prediction \( \mathbf{F}_c \) are compressed into parameters to incorporate historical prediction information; 2) The updated $f_\text{{PML}}$ generates refined classification results \( \{s_f^{\textrm{r}}, s_b^{\textrm{r}}\} \) and boundary offsets \( \{\Delta l^{\textrm{r}}, \Delta o^{\textrm{r}}\} \) based on \( \mathbf{F}_c \). Only anchors with \( s_f > \theta \) (a predefined threshold) are selected, and their boundaries are calculated using Eqn.~\eqref{sted}.

\subsection{Unified Multi-modal Training and Inference}\label{Sec:Unified_training}

We empirically found that directly training a model with hybrid-modal data does not consistently yield strong performance across query types. While models perform well on text queries, performance significantly drops when text is absent (see Fig.~\ref{fig.3_dataset}). To address this, we propose a training strategy called hybrid distillation: \textbf{1)} We train using three query types (text, vision, and vision+text), alternating between them in batches. \textbf{2)} We apply distillation by first training an expert teacher model on text+segment-g queries, which provide the best multi-modal information. This expert model then guides the unified student model through distillation, applied to classification (\(\mathbf{c} = \{ s_f, s_b \}\)), regression (\(\mathbf{r} = \{ \Delta l, \Delta o \}\)), and anchor features (\(\mathbf{F}_{a,i}^{t}\)) with the following loss function: 
\begin{equation}
\begin{aligned}
\mathcal{L}_d = \, \frac{1}{N}\sum_{i=1}^{N}(\mathcal{L}_{\text{KL}}(\mathbf{F}_{a,i}^{s}, \mathbf{F}_{a,i}^{t})  + \mathcal{L}_2(\mathbf{r}_{i}^{s}, \mathbf{r}_{i}^{t})  + \mathcal{L}_2(\mathbf{c}_{i}^{s}, \mathbf{c}_{i}^{t})),
\end{aligned}
\end{equation}
where \(\mathcal{L}_{\text{KL}}\) is KL Divergence and  $\mathcal{L}_2$ is MSE loss, with \(N\) as the number of anchors, and \(s\) and \(t\) as the student and teacher outputs, respectively. Additionally, standard video grounding loss functions are applied to train the student model. The classification head's training loss is defined as:
    \begin{equation}
    \mathcal{L}_{cls} = \frac{1}{N}\sum_{i=1}^{N} \mathcal{L}_{\text{Focal}}(\mathbf{r}_{i}, \mathbf{\hat{r}}_{i}),
    \end{equation}
where we use the Focal loss \cite{ross2017focal} as \(\mathcal{L}_{\text{Focal}}\). The training loss function for the regression head is defined as :
\begin{equation}
\mathcal{L}_{reg} = \frac{1}{N}\sum_{i=1}^{N} (\mathcal{L}_1(\mathbf{\Delta o}_i,\mathbf{\Delta \hat{o}}_i)+\mathcal{L}_1(\mathbf{\Delta l}_i,\mathbf{\Delta \hat{l}}_i)),
\end{equation}
where \(\mathcal{L}_1\) is the L1 loss. The overall loss is defined as:
\begin{equation}
\mathcal{L} = \mathcal{L}_d+\lambda \mathcal{L}_{cls}+\mathcal{L}_{reg},
\end{equation}
where \(\lambda\) is a hyperparameter, and we have found that \(\lambda = 10\) works well across all experiments.

\noindent \textbf{Dynamic Inference Details.} During inference, unlike prior video grounding methods that keep the learned neural network fixed, our model's parameters (\ie, $f_\text{PML}$) are dynamically updated based on the self-supervised loss in Eqn. (\ref{eqn:gml}), allowing it to ``memorize" and leverage historical information to adapt more effectively to unseen data. 

\section{Benchmark Creation and Evaluation Metric}
\label{sec:Dataset}
We establish a new QVHighlights-Unify by expanding the QVHighlights dataset~\cite{lei2021detecting} with image and segment queries. 
\subsection{QVHighlights Dataset}
It covers daily vlogs and news events for both moment retrieval and highlight detection. It contains more than 10,000 videos annotated with free-form queries. Each query is associated with one or multiple variable-length moments in its corresponding video, and a comprehensive 5-point Likert-scale saliency annotation for each clip in the moments.
\subsection{Our QVHighlights-Unify Dataset}
The QVHighlights dataset includes only text queries. We expand it with the following three types of queries.

\noindent\textbf{1) Image-R: retrieved images based on text query.}
To simulate users searching online for visual clues, we first use QVHighlights text queries to retrieve ten semantically matching images. Then, we apply the InternVL vision-language model~\cite{chen2024internvl} to compute similarity scores and select the top-scoring image as the retrieved query. We did not consider retrieving videos because, compared to images, it is substantially more challenging to find a video that accurately matches the text without including irrelevant content.

\noindent\textbf{2) Text-C+Image-C: complementary text-image pairs.}
As noted in \cite{zhang2024localizing}, users may struggle to express unfamiliar or abstract concepts verbally or to find an image that perfectly matches their interests. Providing a simple sketch or sample image alongside a text query can help, as both complement each other semantically to convey the user’s intent. Following \cite{zhang2024localizing}, we modify the text query and generate a complementary image (Image-C) based on the revised text. We also create a corresponding textual description reflecting these modifications (for example, changing “Swimming” to “Dancing” yields “The action is swimming, not dancing.”). Please refer to \cite{zhang2024localizing} for more details.

\noindent\textbf{3) Image/Segment-G: generated visual queries w.r.t. text query.} 
In practical applications, a visual query may not always be retrievable from the internet using its corresponding text query. To address this, we leverage modern generative models to produce images and videos as visual queries. Following \cite{zhang2024localizing}, we design four prompt templates reflecting distinct image styles, randomly pair each text query in the QVHighlights dataset with one template, and use Stable Diffusion~\cite{rombach2022high} for image generation. For videos, we employ the text-to-video model CogVideoX-5B~\cite{yang2024cogvideox} to create a six-second clip per text query as a generated segment query. We then manually filter out visually unclear or semantically mismatched samples, iteratively adjusting the textual input until the output meets the desired criteria.

\subsection{Evaluation Metrics for Online VG}

In online settings, where early and continuous predictions are essential, traditional metrics like mAP fail to account for timeliness. This leads to high scores even when predictions are delayed, making them unrealistic for real-time applications.
To bridge this gap, we introduce two evaluation metrics (\ie, oR@$n$, IoU=$m$ and omAP) that enalize delayed responses by incorporating a decay factor \( \beta \) (\( 0 < \beta < 1 \)). 
If a prediction is made on the ground truth's end time, \( \beta = 1 \); otherwise, \( \beta\) linearly decreases until it reaches zero once the prediction time exceeds the ground truth by a threshold \(t_s \in \{1\text{s}, 3\text{s}, 5\text{s}\}\). 
Although other decay schemes exist (\eg, ~\cite{yuan2021closer}), we adopt linear decay for simplicity. Lastly, we average over these 
\(t_s\) thresholds to obtain the final metrics.

\textbf{1) oR@$n$, IoU=$m$ (oR$_{m}^n$).}  
We extend the standard R@$n$, IoU=$m$ metric by introducing the decay factor \( \beta \). If at least one of the top 
\( n \) retrieved moments have an IoU exceeding \( m \), we set $r(n,m,q_i) = 1$; otherwise, $r(n,m,q_i) = 0$. For moments that match the $i$-th ground truth, we compute \( \beta_i \) using the method above. Formally, we compute
\begin{equation}
\text{oR@}n,\text{IoU@}m = \frac{1}{N_q} \sum_{i=1}^{N_q} \beta_i \cdot r(n, m, q_i),
\end{equation}
where \( N_q \) is the number of queries.

\textbf{2) omAP$_m$.} We define omAP$_m$ as
\begin{align}
\text{omAP$_m$} &= \frac{1}{N_q} \sum_{i=1}^{N_q} \text{oAP$_m^{(i)}$},\\
\text{oAP$_m^{(i)}$} &= \sum_{j=2}^{H_{i}}(\beta_{i,j}
R_{i,j}-\beta_{i,j-1} R_{i,j-1})\beta_{i,j} P_{i,j},
\end{align}  
where \( H_{i} \) is the number of predictions that hit the ground truth corresponding to the $i$-th query, \( P_{i,j} \) and \( R_{i,j} \) are the precision-recall pairs obtained at different cutoff values during Average Precision (AP) calculation, \( \beta_{i,j} \) is the sum of \( \beta \) values for the true positives used in the calculation of \( R_{i,j} \) and \( P_{i,j} \). We multiply \( R_{i,j} \) and \( P_{i,j} \) by \( \beta_{i,j} \) to measure timeliness. Source code will be released.
\section{Main Experiments}
We compare our method with state-of-the-art methods on 4 datasets, including our QVHighlights-Unify dataset, ANet-Captions~\cite{caba2015activitynet}, TACoS~\cite{rohrbach2014coherent}, and MAD~\cite{soldan2022mad} datasets.
Due to page limits, we put more details in the supplementary.

\subsection{Results on the QVHighlights-Unify Dataset} 

\begin{figure}[t]

    \includegraphics[width=\columnwidth]{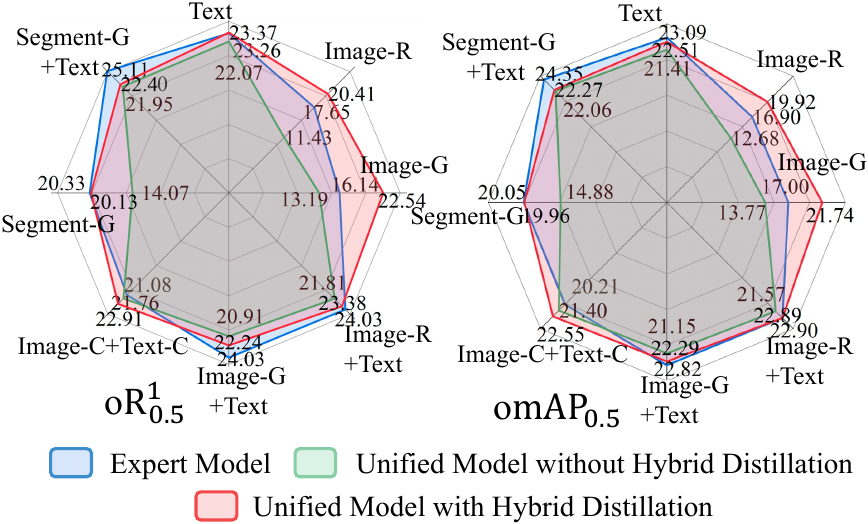}
    \vspace{-20pt}
    \caption{Performance comparisons on our QVHighlights-Unify dataset w.r.t. oR$_{0.5}^1$ and omAP$_{0.5}$.}
    
    \vspace{-9pt}
    \label{fig.3_dataset}
\end{figure}

The hybrid-modal query framework includes eight distinct input configurations, incorporating both single-modal and dual-modal queries. Figure \ref{fig.3_dataset} compares three models: 1) an expert model trained for each query type (blue), 2) a unified model directly trained for hybrid-modal queries (green), and 3) a unified model trained with hybrid distillation (red).

\noindent\textbf{Training an expert model for each query type.} 
We observe \textit{1) Segment queries outperform Image queries:} This difference (20.33\% vs 16.14\%) likely arises as video grounding retrieves dynamic video segments, and Segment queries are better suited than Image queries to accurately describe a user's content of interest.
\textit{2) Multimodal queries outperform single-modal queries:} For example, the Segment-G + Text expert model achieves a performance metric of 25.11\%, substantially higher than either the Segment-G (20.33\%) or Text (23.26\%). This result suggests that multimodal queries provide richer and more comprehensive information about the desired moment.

\noindent\textbf{Training a unified model to handle all query types.} Our key findings include \textit{1) Challenges with training a hybrid-modal unified model:} When multiple query types are directly combined into a single unified model, performance generally declines compared with the expert model. As shown in Figure~\ref{fig.3_dataset}, visual queries (e.g., Segment-G 14.07\%) are considerably lower than the text query's score (22.07\%). This suggests that during training, the model tends to prioritize the dominant modality, suppressing the optimization of other modalities~\cite{zhou2023intra}.
\textit{2) Improvement with hybrid distillation:} The model's performance improved significantly, especially in cases without text query. Specifically, when only an Image-R query is used, the oR$_{0.5}^1$ metric increases by 8.98\% (from 11.43\% to 20.41\%), demonstrating the effectiveness of our proposed approach.

\noindent\textbf{Comparisons with other VG methods.} 
Following the implementation in~\cite{gan2023temporal}, we adapt SoTA offline video grounding (VG) algorithms for the online VG task. Additionally, we re-implement TwinNet on our dataset. Notably, all compared methods utilize CLIP features for both video and text modalities. Given that previous approaches exclusively employ text queries during training, we conduct evaluations on the QVHighlights-Unify benchmark with text query as input. The detailed results are shown in Table \ref{Tab:QVHighlights}. Our method exhibits notable improvements across various metrics.

\begin{table}[t]
\centering
\scriptsize
\caption{Comparisons with SoTA models on QVHighlights-Unify.}
\vspace{-8pt}
\resizebox{0.7\columnwidth}{!}{
\begin{tabular}{cccc}
\hline
Setting (Text Query) & Method & oR$_{0.5}^1$ & omAP$_{0.5}$ \\ \hline
\multirow{3}{*}{\begin{tabular}[c]{@{}c@{}}Offline VG\\ (Modified to online)\end{tabular}} & TaskWeave~\cite{yang2024task} & 7.02 & 5.96 \\
 & TR-DETR~\cite{sun2024tr} & 7.37 & 6.06 \\
& R$^2$-Tuning~\cite{liu2025mathrm} & 9.30 & 8.17 \\ \hline
\multirow{2}{*}{Online VG} & TwinNet~\cite{ferriol2022building} & 20.78 & 19.73 \\
 & Ours & \textbf{23.26} & \textbf{23.09} \\ \hline
\end{tabular}
}
\label{Tab:QVHighlights}
\vspace{-5pt}
\end{table}

\subsection{Results on Text Query-based VG Datasets}
For a more comprehensive comparison, we evaluate our method against baselines on existing text query-based datasets. To ensure fairness, we employ the ANet-Captions and TACoS datasets with C3D features, and the MAD dataset with CLIP features. Following \cite{gan2023temporal}, we not only compare variants of offline VG modified for online settings but also evaluate several online action detection methods (likewise modified for online VG). These baseline results are directly provided by \cite{gan2023temporal}. Since we do not have access to the models and therefore cannot measure the online metrics, we compare the offline metrics to ensure fairness and consistency. As shown in Table \ref{Tab:comparison}, our method substantially outperforms TwinNet and other approaches. Specifically, for \(R_{0.7}^1\), our method achieves an improvement of 1.80\% over TwinNet on ANet-Captions. These findings further underscore the unique challenges presented by online VG compared to offline VG, indicating the need for specialized strategies to address these challenges.

\begin{table}[t]
\centering
\caption{Results on ANet-Captions, TACoS, and MAD datasets.}
\vspace{-8pt}
\resizebox{\columnwidth}{!}{
\begin{tabular}{cc|cc|cc|cc}
\hline
\multirow{2}{*}{Setting} & \multirow{2}{*}{Method} & \multicolumn{2}{c|}{ANet-Captions} & \multicolumn{2}{c|}{TACoS} & \multicolumn{2}{c}{MAD} \\
 &  & R$_{0.5}^1$ & R$_{0.7}^1$ & R$_{0.5}^1$ & R$_{0.7}^1$ & R$_{0.3}^5$ & R$_{0.5}^5$ \\ \hline
\multirow{3}{*}{\begin{tabular}[c]{@{}c@{}}Online \\Action Detection\\ (Modified  to VG)\end{tabular}} & OadTR~\cite{wang2021oadtr} & 23.27 & 10.97 & 21.12 & 10.92 & 2.50 & 0.90 \\ 
 & LSTR~\cite{xu2021long} & 24.05 & 11.19 & 26.02 & 16.75 & 3.56 & 1.43 \\
 & GateHUB~\cite{chen2022gatehub} & 23.30 & 11.31 & 27.10 & 17.25 & 3.38 & 1.47 \\ \hline
\multirow{7}{*}{\begin{tabular}[c]{@{}c@{}}Offline VG\\ (Modified to online)\end{tabular}}
 & VSLNet~\cite{zhang2020span} & 12.89 & 5.05 & 25.74 & 12.60 & - & - \\
 & 2DTAN~\cite{zhang2020learning} & 8.39 & 2.96 & 6.82 & 3.32 & - & - \\
 & SeqPAN~\cite{zhang2021parallel} & 12.57 & 4.79 & 25.07 & 13.67 & - & - \\
 & SMIN~\cite{wang2021structured} & 7.47 & 2.64 & 6.00 & 2.92 & - & - \\ 
 & TaskWeave~\cite{yang2024task} & 8.22 & 3.67 & 14.93 & 6.78 & - & - \\
 & TR-DETR~\cite{sun2024tr} & 10.37 & 4.31 & 16.25 & 7.44 & - & - \\
 & R$^2$-Tuning~\cite{liu2025mathrm} & 9.17 & 4.16 & 21.69 & 11.24 & - & - \\ \hline
\multirow{2}{*}{Online VG} & TwinNet~\cite{ferriol2022building} & 25.48 & 12.56 & 29.74 & 19.07 & 4.71 & 2.00 \\
 & Ours & \textbf{26.57} & \textbf{14.36} & \textbf{30.98} & \textbf{21.17} & \textbf{6.32} & \textbf{3.27} \\ \hline
\end{tabular}
}
\label{Tab:comparison}
\vspace{-5pt}
\end{table}

\begin{table}[t]
\centering
\caption{Computational overhead of PMB and dynamic updates.}
\vspace{-8pt}
\tiny
\resizebox{\columnwidth}{!}{
\begin{tabular}{cccccc}
\hline
Method & Latency(ms) & FPS & FLOPs(M)& MACs(M) \\ \hline
Overall Model& 21.76&45.95&5932.42&2966.21\\
PMB& 2.20& 454.54& 11.43 &5.72 \\
Dynamic Update& 0.30 &3333.30&1.17&0.59\\ \hline
\end{tabular}
}
\label{tab:efficiency}
\vspace{-15pt}
\end{table}

\subsection{Runtime Analysis of Our Method}
As we focus on the online VG problem, runtime efficiency is critical. To evaluate its performance, we test the model on a single RTX 4090 GPU.
All metrics, including latency, FPS, FLOPs, and MACs, are computed on a per-frame basis. 
As shown in Table \ref{tab:efficiency}, the overall model achieves an FPS of 45.95, satisfying real-time processing requirements. Moreover, the FLOPs and latency of the PMB and Dynamic Update components are significantly lower than those of the entire model, indicating that both the proposed PMB and the dynamic update process exhibit high efficiency.
\section{Ablation Studies}
\begin{table}[t]
\centering
\caption{\textbf{Left:} Effect of parametric memory layer. $f_{\text{PML}}$ is replaced by LSTM and self-attention (ATT), respectively. \textbf{Right:} Ablation study on the inputs of prediction refinement module: w/o Refine (no prediction refinement module), Pred (prediction only), and Pred+AF (prediction with anchor features).}
\vspace{-4pt}
\resizebox{\linewidth}{!}{
\small
\begin{tabular}{llcc|lcc}
\hline
Query & Variant   & oR$_{0.5}^1$   & omAP$_{0.5}$ & Variant   & oR$_{0.5}^1$   & omAP$_{0.5}$ \\ \hline
\multirow{3}{*}{Text} & Ours-ATT   & 13.93  & 16.41 & w/o Refine   &  17.64   &  17.43 \\ 
& Ours-LSTM   & 22.37 & 21.66 & Pred   &  18.99     &  21.07 \\
& Ours   & \textbf{23.37} & \textbf{22.51}  & Pred+AF   & \textbf{23.37} & \textbf{22.51}  \\ 
\hline
\multirow{3}{*}{Image-R} & Ours-ATT   & 10.71 & 13.3 & w/o Refine   & 16.20 & 15.95\\ 
& Ours-LSTM   & 18.96 & 18.92 & Pred   & 16.66 & 18.68 \\
& Ours  &   \textbf{20.41}  & \textbf{19.92} & Pred+AF  &  \textbf{20.41}  & \textbf{19.92} \\ 
\hline
\multirow{3}{*}{Image-G} & Ours-ATT & 11.69 & 14.35 & w/o Refine & 16.89  & 17.00 \\ 
& Ours-LSTM  & 19.26 & 18.69 & Pred  &  18.96  & 20.82 \\
 & Ours & \textbf{22.54} & \textbf{21.74}  & Pred+AF  & \textbf{22.54} & \textbf{21.74} \\ 
\hline
\multirow{3}{*}{Image-R+Text} & Ours-ATT   & 13.55 & 15.89 & w/o Refine   &   17.23  & 17.69 \\ 
& Ours-LSTM   & 22.39 & 21.41 & Pred   &   18.78    & 20.96 \\
& Ours  &  \textbf{23.38} & \textbf{22.89} & Pred+AF  &  \textbf{23.38} & \textbf{22.89} \\ 
\hline
\multirow{3}{*}{Image-G+Text} & Ours-ATT   & 14.42 & 16.17 & w/o Refine   &  18.90   & 18.61 \\ 
& Ours-LSTM   & 21.97 & 21.46 & Pred   & 20.05 & 21.89 \\
& Ours & \textbf{22.24} & \textbf{22.29} & Pred+AF  & \textbf{22.24} & \textbf{22.29} \\   
\hline
\multirow{3}{*}{Image-C+Text-C} & Ours-ATT   & 12.2 & 15.33 & w/o Refine   & 16.53    & 17.02 \\ 
& Ours-LSTM   & 20.46 & 20.24 & Pred   &   19.61   &  21.30 \\
& Ours   & \textbf{22.91} & \textbf{22.55} & Pred+AF   & \textbf{22.91} & \textbf{22.55} \\ 
\hline
\multirow{3}{*}{Segment-G} & Ours-ATT & 11.85 & 14.2 & w/o Refine & 15.43 &  15.88 \\ 
& Ours-LSTM   & 17.41 & 16.93 & Pred   & 17.30   & 19.28 \\
& Ours  &  \textbf{20.13} & \textbf{19.96} & Pred+AF  &  \textbf{20.13} & \textbf{19.96} \\ 
\hline
\multirow{3}{*}{Segment-G+Text} & Ours-ATT   & 13.32 & 15.48 & w/o Refine   & 17.69   & 18.12 \\ 
& Ours-LSTM   & 21.95 & 21.14 & Pred   &   19.76    & 21.39 \\
& Ours  &  \textbf{22.40}   & \textbf{22.27}  & Pred+AF  &  \textbf{22.40}   & \textbf{22.27} \\ 
\hline
\label{Tab:structure}
\end{tabular}
}
\vspace{-10pt}
\end{table}

\begin{figure}[t]
    \includegraphics[width=\columnwidth]{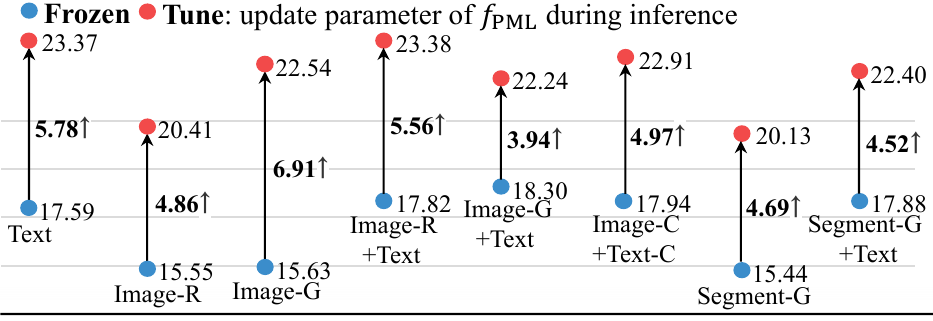}
    \vspace{-20pt}
    \caption{Effectiveness of test-time model updates w.r.t. oR$_{0.5}^1$.}
    \label{fig.4_test_time}
    \vspace{-12pt}
\end{figure}

\subsection{\texorpdfstring{Does $f_{\text{PML}}$ Help Online Video Grounding?}{Does fPML Help Online Video Grounding?}}
In our approach, both feature fusion and prediction refinement are embedded within the proposed $f_{\text{PML}}$. We designed two variants for comparison: \textbf{1) Ours-LSTM:} where $f_{\text{PML}}$ is replaced with an LSTM, and \textbf{2) Ours-ATT}: where $f_{\text{PML}}$ is replaced with a self-attention layer of equivalent parameter size. The remaining network structures are identical to our method to ensure a fair comparison. As shown in Table \ref{Tab:structure}, both the LSTM and our $f_{\text{PML}}$ consistently outperform the self-attention (ATT) layer, with $f_{\text{PML}}$ achieving 23.37\% compared to 13.93\% for the ATT layer, highlighting the importance of incorporating historical information in online video grounding task. Furthermore, across different query configurations, our method surpasses the LSTM in all cases, notably improving the Text query from 22.37\% to 23.37\% and the Segment-G query from 17.41\% to 20.13\%, further highlighting that when modeling historical information, a more expressive neural network—such as $f_{\text{PML}}$—is superior to a fixed-size hidden state, as it provides more effective information for current predictions.

\subsection{What Benefits Prediction Refinement?}
In our prediction refinement module, the $f_{\text{PML}}$ parameter encapsulates both the current prediction and the current anchor feature (AF), compressing the information of the current step. This approach models the historical context of both the prediction and the anchor feature. We progressively removed these two types of information and present the results in Table \ref{Tab:structure}. Removing the anchor feature input leads to a significant decline (from 23.37\% to 18.99\%) in the oR$_{0.5}^1$ metric when using a text query. Moreover, when the entire Prediction Refinement Head is eliminated, the performance deteriorates (1.35\%) even further. These results highlight the critical role of prediction information memory and demonstrate that including anchor features considerably improves model performance.

\subsection{\texorpdfstring{Does Updating $f_{\text{PML}}$ in Inference Time Help?}{Does Updating fPML in Inference Time Help?}}

The key feature of our method is that, upon the arrival of each new video, the parameters of our model (i.e., \( f_\text{PML} \)) are reset and dynamically updated with each frame input, based on the self-supervised loss defined in Eqn. (\ref{eqn:gml}). To investigate the impact of this strategy on online video grounding performance, we compare two implementation variants: 1) \textbf{Frozen}: Parameters of \( f_\text{PML} \) are kept fixed during inference.  2) \textbf{Tune}: the parameters of \( f_\text{PML} \) are dynamically updated during inference. As shown in Figure~\ref{fig.4_test_time}, the Tune configuration consistently outperforms Frozen across eight distinct query composition settings. These results indicate that updating \( f_\text{PML} \) during the testing phase enables better adaptation to unseen data.
\section{Conclusion}
We have introduced Online Video Grounding with Hybrid-modal Queries (OVG-HQ), extending traditional video grounding task to support text, images, video snippets, and their combinations in streaming scenarios.
To enable this, we have developed QVHighlight-Unify and introduced two new metrics to jointly evaluate accuracy and timeliness.
To benchmark OVG-HQ, we have proposed OVG-HQ-Unify, a unified model featuring a Parametric Memory Block for retaining past context and a hybrid-distillation strategy for training.
We hope this work inspires further research in online video grounding, bridging the gap between academic benchmarks and real-world applications.
{\flushleft \bf Acknowledgements}. This work was partially supported by National Natural Science Foundation of China (NSFC) under Grants 62202311, the Guangdong Basic and Applied Basic Research Foundation under Grants 2023A1515011512, Key Scientific Research Project of the Department of Education of Guangdong Province 2024ZDZX3012.
{
    \small
    \bibliographystyle{ieeenat_fullname}
    \bibliography{main}
}
\clearpage
\setcounter{page}{1}
\setcounter{section}{0}
\renewcommand\thesection{\Alph{section}}
\setcounter{figure}{0}
\renewcommand\thefigure{\Alph{figure}}
\setcounter{table}{0}
\renewcommand\thetable{\Alph{table}}
% \title{OVG-HQ: Online Video Grounding with Hybrid-modal Queries \\
% Paper ID 7351}
\maketitlesupplementary
\captionsetup{font={normalsize}}
\captionsetup[subfigure]{font=normalsize}

In the supplementary material, we provide more details and more experimental results of our work. We organize the supplementary into the following sections.

\begin{itemize}
    \item In Section~\ref{Sec:model_details}, we present a detailed description of our proposed OVG-HQ-Unify framework, which consists of the feature extraction, memory-guided multi-modal fusion module, memory-guided moment prediction module.

    \item In Section~\ref{Sec:more_ablation}, we provide results on ICQ-Highlight dataset and results of the ablation experiments.

    \item In Section~\ref{Sec:data_details}, we provide a comprehensive description of the process used to construct our hybrid-modal queries dataset, QVHighlights-Unify. Additionally, we showcase some examples of the samples we generate.

    \item In Section~\ref{Sec:reliability}, we analyze the diversity, realism, and construction reliability of our dataset, and provide a statistical comparison with existing benchmarks.

    \item In Section~\ref{Sec:more_exp}, we present additional experimental results, including the utility of multimodal queries for existing offline grounding methods.
\end{itemize}

\begin{figure*}[t] \includegraphics[width=\textwidth]{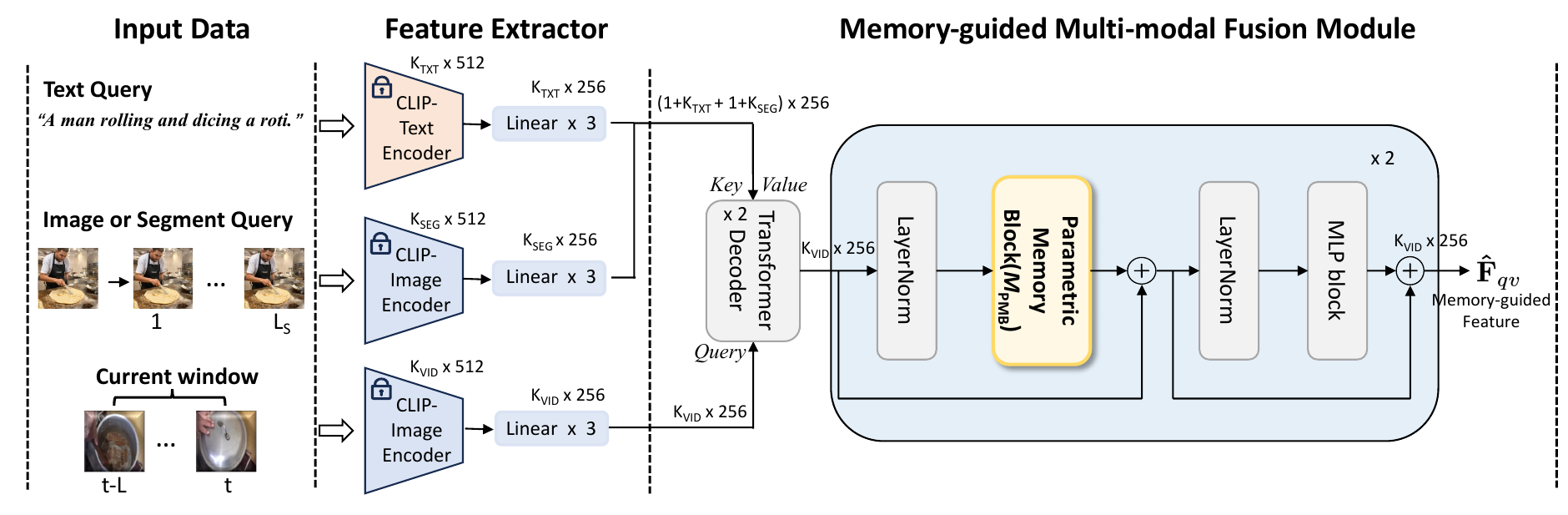} \caption{Detailed illustration of the Input Data, Feature Extractor, and Memory-guided Multi-modal Fusion Module. ``Linear $\times$ N" indicates N consecutive linear layers coupled with Layer Normalization.} \label{Fig.sup_memoryguided} \end{figure*}

\begin{figure*}[t]
    \includegraphics[width=\textwidth]{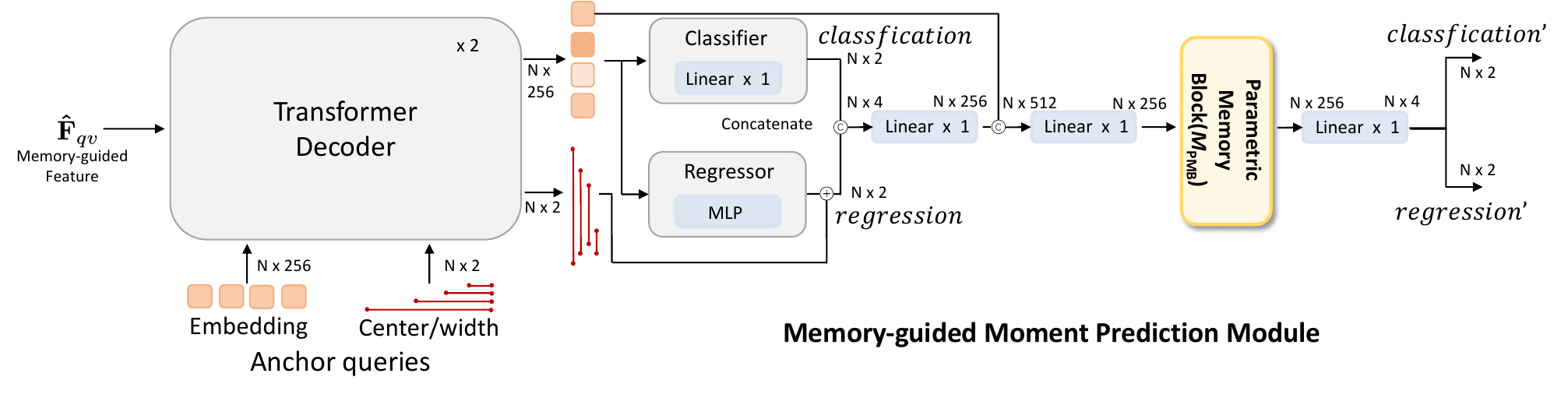}
    \caption{Detailed illustration of Memory-guided Moment Prediction Module.} 
    \label{Fig.sup_refinement}
\end{figure*}

\section{More Details of Model and Implementation}\label{Sec:model_details}
We provide the implementation details in this section. We will release the collected dataset, source code and trained models upon acceptance.
\subsection{Details of Feature Extraction}
This section introduces the feature extraction process, as shown in Figure~\ref{Fig.sup_memoryguided}. The feature extraction module accepts arbitrary combinations of multi-modal queries and current video window as  input, and outputs query features and video features via the query feature extractor and the video feature extractor. When a query includes multiple modalities, learnable modality tokens are inserted at the beginning of each query feature sequence to enable the model to distinguish different modalities. Below, we separately introduce the video feature extractor and the query feature extractor we employ.
\subsubsection{Video Feature Extractor}
We process streaming video through a sliding window of size \( L \) seconds advancing at \( M=2 \)-second intervals. At each time step \( t \), the CLIP image encoder~\cite{radford2021learning} extracts features from current frame, with overlapping window features computed once and cached. For initial windows containing fewer than \( K \) frames, we left-pad zero vectors to maintain the feature matrix dimensionality \( \mathbf{F}_v \in \mathbb{R}^{K \times D_v} \).

\subsubsection{Query Feature Extractor} 
\textbf{1) Text Query}: The CLIP text encoder~\cite{radford2021learning} generates textual features \( \mathbf{F}_t\). \\
\textbf{2) Segment Query}: Employing the video feature extractor with 2-second sampling intervals to obtain \( \mathbf{F}_s \). \\
\textbf{3) Image Query}: The CLIP image encoder~\cite{radford2021learning} extracts features \( \mathbf{F}_i \), which are duplicated temporally to match the segment query length.
\subsection{Memory-guided Multi-modal Fusion Module}    

We employ a two-layer Transformer Decoder~\cite{moon2023query} to fuse the video features and query features, resulting in query-aware video representations \( \mathbf{F}_{qv} \in \mathbb{R}^{K_{\text{VID}} \times D} \). Subsequently, we input \( \mathbf{F}_{qv}\) into two layers of residual block composed of parametric memory blocks, integrating historical information to obtain memory-guided features \( \mathbf{\hat{F}}_{qv} \in \mathbb{R}^{K_{\text{VID}} \times D}\). We set $K_{\text{VID}}$=16. The detailed process is illustrated in Figure~\ref{Fig.sup_memoryguided}.

\subsection{Memory-guided Moment Prediction Module}
We employ a two-layer Transformer Decoder~\cite{moon2023query} to decode the memory-guided features. \textbf{First}, the Anchor Embeddings and the Anchors' center and width coordinates are used as queries for the Decoder. Through this decoding process, we obtain the anchor features and the refined anchor center and width coordinates. \textbf{Second}, the anchor features produced by the Transformer Decoder are input into a classifier and a regressor to obtain classification and regression results. \textbf{Third}, these results are passed to subsequent parametric modules, where they are combined with historical information to generate optimized predictions. We employ 4 anchor queries with lengths of 1, 2, 4, 8.  The detailed procedure is illustrated in Figure~\ref{Fig.sup_refinement}. 

\subsection{Implementation Details}
We use PyTorch 2.0.0 and 1 4090 GPUs for our experiments. The model weights are initialized using Xavier initialization~\cite{glorot2010understanding}. We optimize the model parameters using the AdamW~\cite{loshchilov2017fixing} optimizer, with an initial learning rate of 1e-4 and a weight decay of 1e-4. The model is trained for 30 epochs; The batch size is set to 256. In the focal loss~\cite{ross2017focal}, we set \(\alpha = 0.9\) and \(\gamma = 2\), and apply a dropout rate of 0.5. In the distillation process, the temperature for the KL divergence is set to 2. During training, only anchors with a temporal Intersection over Union (tIoU) greater than 0.5 with the ground truth are considered training samples. 

\section{More Ablation Study Results}\label{Sec:more_ablation}

\subsection{More results on ICQ-Highlight dataset and Comparisons with hybrid-modal retrieval methods.}
To further validate effectiveness of our approach, we conducted comprehensive experiments on the ICQ-Highlight~\cite{zhang2024localizing} dataset containing abundant complementary image-text query pairs. Following the summarization paradigm proposed in~\cite{zhang2024localizing}, we employ a MLLM (LLaVA-mistral-1.6)~\cite{liu2023visual, liu2024improved} to convert paired image-text inputs into unified textual queries. As demonstrated in Table \ref{tab:icq_highlight}, when feeding these synthesized textual queries into models pre-trained on QVHighlights, our method significantly outperforms the TwinNet baseline across all four image style categories in evaluation metrics, thereby conclusively validating the efficacy of our model design. Notablly, our proposed QVH-HQ-Unify framework achieves state-of-the-art performance across all four image style when directly processing raw multimodal queries from ICQ-Highlight. This demonstrates the remarkable advantages of our unified framework in cross-modal semantic fusion, which effectively captures complementary semantic information from heterogeneous image-text queries.

\begin{table}[ht]
\centering
\caption{Comparison between our model and TwinNet across four distinct styles.}
\resizebox{\columnwidth}{!}{
\begin{tabular}{ccccc}
\hline
Method & scribble & cartoon& cinematic& realistic \\ \hline
TwinNet-MLLM & 18.97   & 18.62 & 17.45   & 18.37\\
Ours-MLLM    & 19.95 & 20.34   & 19.53 & 19.08 \\ 
Ours-QVH-HQ-Unify &21.98  & 22.35 & 21.25  & 22.07\\ \hline
\end{tabular}
}
\label{tab:icq_highlight}
\end{table}

\section{Dataset Construction Details}\label{Sec:data_details}
Our dataset is constructed based on QVHighlights~\cite{lei2021detecting}, which originally contains only text queries and raw videos. To expand the modalities of queries, we enrich the dataset by incorporating additional types of queries.
\subsection{Detailed construction pipeline}

\begin{figure*}[t] \includegraphics[width=\textwidth]{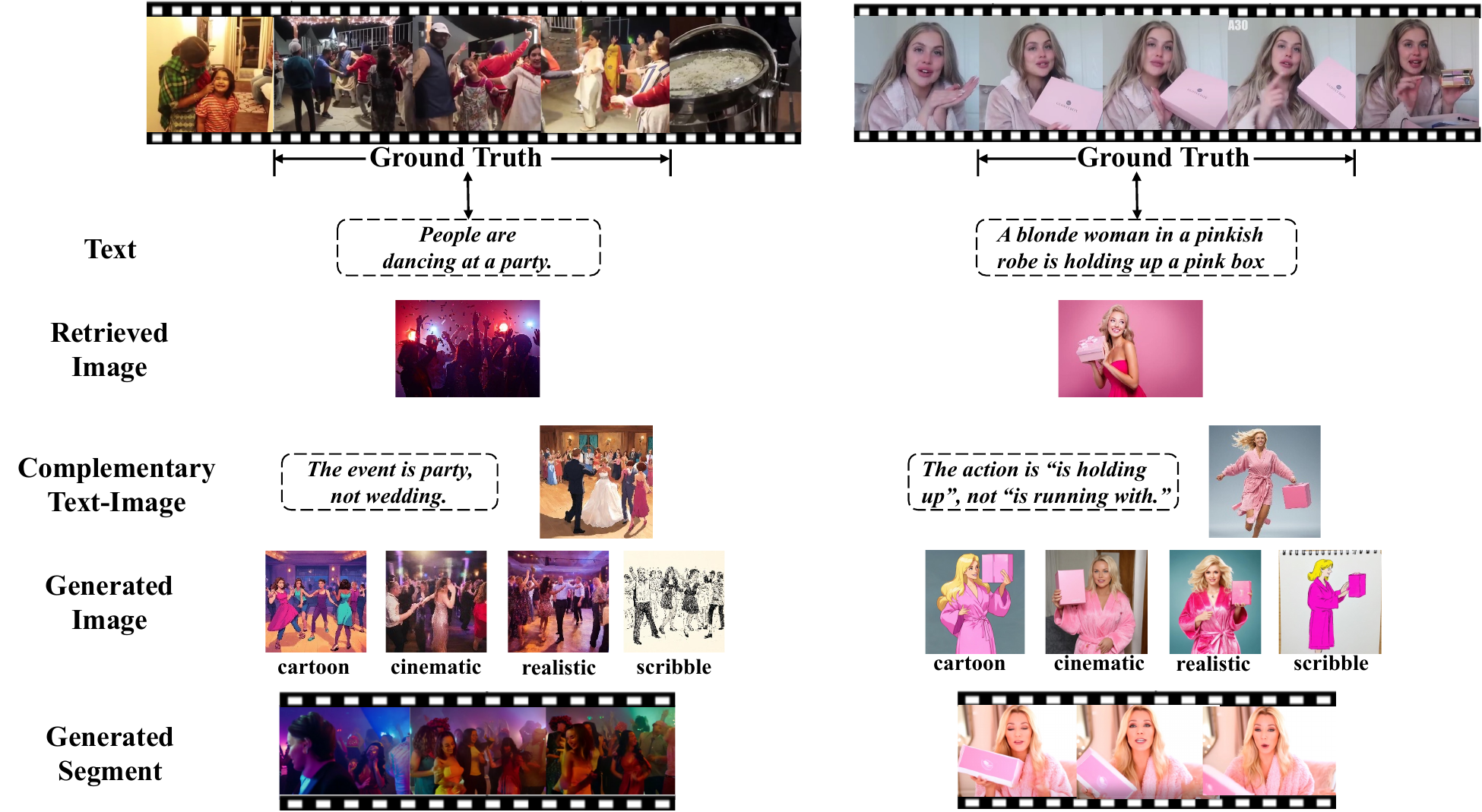}
\caption{Examples of our constructed QVHighlights-Unify dataset.} \label{Fig.sample} 
\end{figure*}

In this section, we present the construction processes of the four types of queries: Retrieved image (Image-R), generated image (Image-G), generated video segment (Segment-G), and complementary text-image (Text-C and Image-C) as shown in Figure~\ref{Fig.sample}:
\begin{itemize}

    \item \textbf{Image-R Query}: Using the original text queries from the QVHighlights dataset, we retrieved ten semantically matching images from the internet. We then employed the advanced vision-language model, InternVL~\cite{chen2024internvl}, to compute the similarity scores between these ten images and the text. The image with the highest similarity score was selected as our final choice.
    \item \textbf{Image-G Query}: Consistent with the method in ICQ-Highlight~\cite{zhang2024localizing}, we generate images in four styles: scribble, cartoon, cinematic, and realistic. We carefully design specific prompts for each style. By combining the text queries from QVHighlights with these prompts, we use the text-to-image generation model Stable Diffusion~\cite{rombach2022high} to produce images in these four different styles.

    \item \textbf{Segment-G Query}: Each text query in the QVHighlights dataset is input into GPT-4o to generate a longer, richer, more detailed text description. This generated text is then fed into the text-to-video model CogVideoX-5B~\cite{yang2024cogvideox} to produce a 6-second video, serving as the generated segment query. \\
    During our review of the generated data, we identify a small number of low-quality generated segments. Specifically, we find that the model occasionally generates meaningless pure white videos, whose file sizes are generally less than 150KB. To address this issue, we first filter out low-quality segments based on their file size, considering any segment smaller than 150KB as low quality. For all such segments, we re-extract the original video frames and generate more detailed, and semantically similar text queries. We continue generating new segments until their file sizes exceed 150KB. Subsequently, we manually review all the segments and, when necessary, adjust the input text for CogVideoX-5B to ensure that the generated segments meet all required standards.

    \item \textbf{Image-C and Text-C}: Drawing inspiration from the data construction methodology of ICQ-Highlight~\cite{zhang2024localizing}, we created Image-C and Text-C pairs through a text modification process. Specifically, the revised text queries (e.g., replacing ``Swimming'' with ``Dancing'') were fed into the ~\cite{rombach2022high} Stable Diffusion model to generate corresponding Image-C, while the modification deltas between original and revised texts were formulated into Text-C (e.g., ``The action is swimming, not dancing''). This procedural alignment ensures that: (1) Image-C visually embodies the semantics of the revised text through diffusion-based synthesis; (2) Text-C textually articulates the modification intent.
\end{itemize}

\begin{figure*}[t]
    \centering
    \begin{subfigure}[t]{0.75\textwidth}
        \centering
        \includegraphics[width=\linewidth]{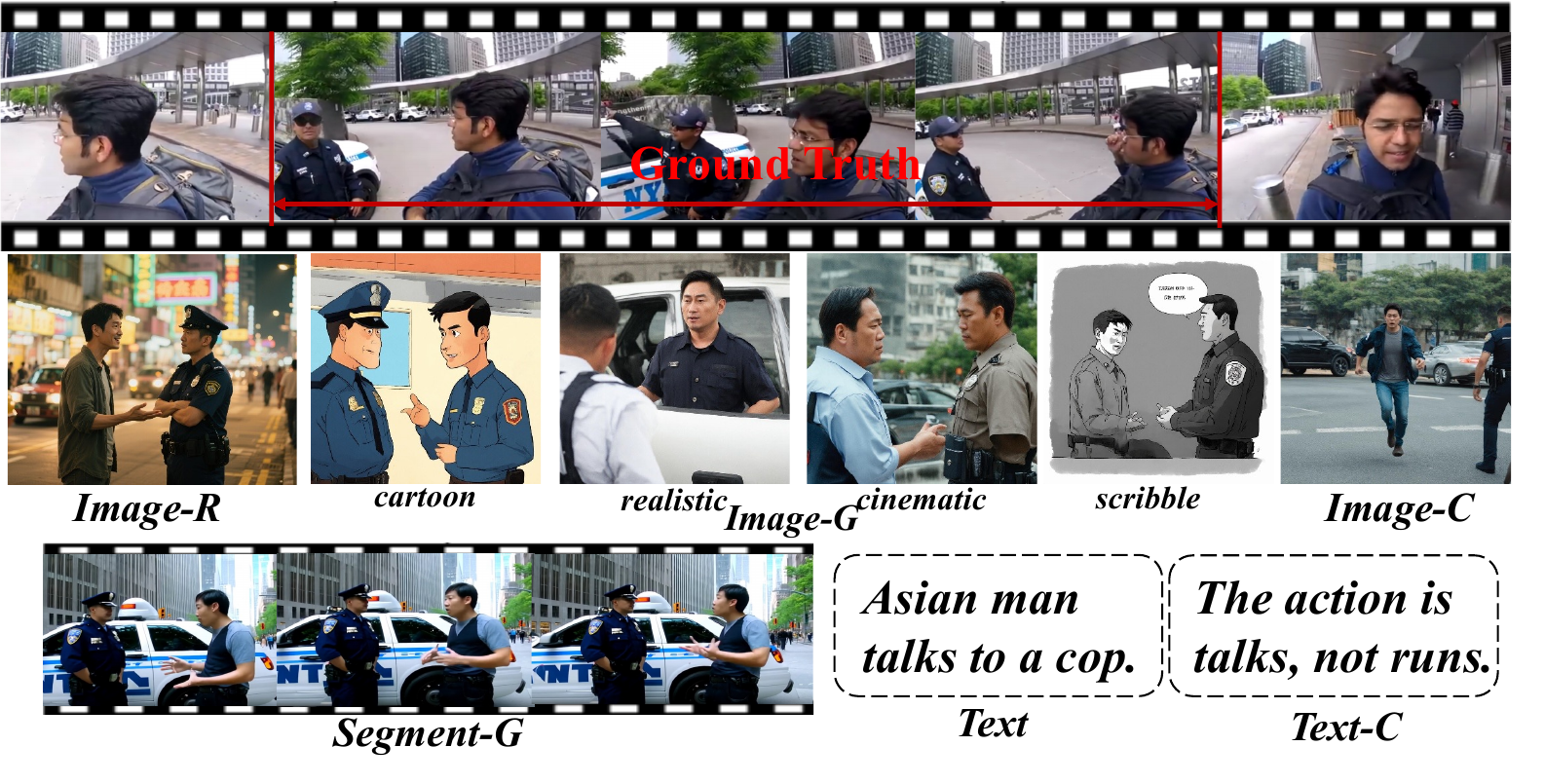}
        \caption{An example of our generated dataset queries.}
        \label{fig:vis_example}
    \end{subfigure}
    \hfill
    \begin{subfigure}[t]{0.23\textwidth}
        \centering
        \includegraphics[width=\linewidth]{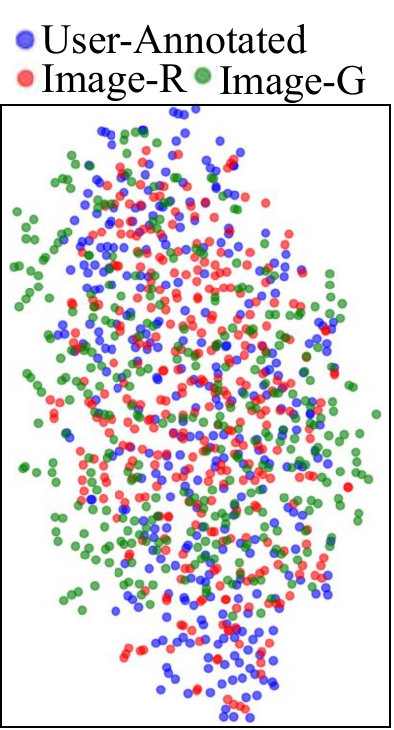}
        \caption{t-SNE of image features.}
        \label{fig:tsne_features}
    \end{subfigure}
    
    \vspace{4pt} 
    \caption{Qualitative and quantitative analysis of our constructed dataset. (a) An example of different query modalities generated for a single event. (b) A t-SNE visualization showing the feature space overlap between ground-truth frames (GT), retrieved images (Image-R), and generated images (Image-G), confirming their semantic alignment.}
    \label{Fig:dataset_analysis}
\end{figure*}

\section{Analysis of Dataset and Query Reliability}\label{Sec:reliability}

\subsection{Diversity and Realism of Multi-modal Queries}
To ensure the realism and diversity of our dataset, the construction process is grounded in genuine human behaviors and interests. The text queries in the base QVHighlights dataset were collected via Amazon Mechanical Turk, where annotators described events they found interesting after watching entire videos. This methodology ensures the queries reflect natural human curiosity and language. Consequently, the multi-modal queries derived from these text descriptions inherit a high degree of realism and diversity, representative of real-world user scenarios.

To further simulate user behavior and mitigate biases in the visual query construction, we employed two distinct pipelines. For the \textbf{retrieval pipeline} (Image-R), we emulate a user searching for a relevant image online. We first perform a broad image search and then use the InternVL model to select the image most semantically aligned with the text query, a process shown to correlate highly with human preference (see Section~\ref{subsec:reliability_internvl}). For the \textbf{generative pipeline} (Image-G), we generate images in four distinct visual styles (scribble, cartoon, cinematic, and realistic) to enhance diversity and promote model generalization, as illustrated in Figure~\ref{Fig:dataset_analysis}.

To empirically validate that our generated and retrieved queries align with real user intent, we used CLIP to extract features for Image-R, Image-G, and a ground-truth frame from the user-annotated video segment for the same text query. A t-SNE visualization of these features, shown in Figure~\ref{Fig:dataset_analysis} (b), reveals a strong overlap in the feature space. This indicates that our constructed visual queries effectively capture diverse user intentions and are semantically consistent with user-annotated ground truth content.

\begin{table}[t]
\centering
\caption{Statistical comparison with existing datasets.}
\resizebox{\columnwidth}{!}{
\begin{tabular}{lccc}
\hline
Dataset       & Text Query & Image Query & Segment Query \\
\hline
QVHighlights   & 10.3K & 0        & 0        \\
ICQ-Highlight & 1.5K       & 6.2K     &  0    \\
Ours          & 19.0K & 26.3K & 8.8K \\
\hline
\end{tabular}}
\label{Tab:dataset_comparison}
\end{table}

\subsection{Reliability of the Dataset Construction Pipeline}\label{subsec:reliability_internvl}
The quality of our dataset relies on the foundational models used in its construction. To validate the reliability of using InternVL for ranking retrieved images (Image-R), we conducted a human evaluation. We randomly selected 200 text queries and had human annotators manually rank the top 10 retrieved image candidates for each query based on semantic similarity. The Pearson correlation coefficient between the manual rankings and the InternVL rankings was 0.86, indicating a very strong positive correlation. Furthermore, the top-3 consistency, where both methods yield the identical top three results in the same order, reached 96\%. These results demonstrate that InternVL's automated ranking aligns closely with human judgment, confirming the method's reliability and scalability for dataset construction. For all generated queries (Image-G and Segment-G), a manual filtering process was also applied to ensure high quality and semantic relevance, as detailed in Section 4.2.

\subsection{Statistical Comparison with Existing Datasets}
Our QVHighlights-Unify dataset significantly expands upon existing benchmarks for video grounding by incorporating a greater number of modalities and a larger volume of query samples. Table~\ref{Tab:dataset_comparison} provides a statistical comparison with the original QVHighlights and the ICQ-Highlight datasets. Our dataset not only includes the original text queries but also introduces multiple new visual query types (retrieved images, generated images) and video segment queries, resulting in a more comprehensive and practical benchmark for the OVG-HQ task.

\section{Additional Experimental Results}\label{Sec:more_exp}

\subsection{Utility of Multimodal Queries in Offline Grounding Methods}
To demonstrate that the benefit of hybrid-modal queries is not limited to our online framework, we adapted three representative offline video grounding methods (Moment-DETR, QD-DETR, TaskWeave) to accept multimodal inputs. We augmented their architectures with image and video encoders alongside their original text encoders and evaluated them on our dataset. As shown in Table~\ref{Tab:offline_multimodal}, providing visual queries (Image-R or Segment-G) in addition to text consistently improves the performance of these established offline methods. This confirms the general utility of using hybrid-modal queries for video temporal grounding tasks.

\begin{table}[t]
\centering
\caption{Effect of multimodal queries on representative offline video grounding methods, evaluated on R$_{0.5}^1$. Adding visual queries (+Image-R or +Segment-G) consistently improves performance over text-only queries.}
\resizebox{\columnwidth}{!}{
\begin{tabular}{lcccc}
\hline
Method & Publication & Text & +Image-R & +Segment-G \\
\hline
Moment-DETR& NeurIPS21 & 53.94 & 55.28 (\textbf{+1.34}) & 55.37 (\textbf{+1.43}) \\
QD-DETR & CVPR23 & 62.68  & 64.12 (\textbf{+1.44}) & 64.03 (\textbf{+1.35}) \\
TaskWeave & CVPR24 & 64.26 & 65.37 (\textbf{+1.11}) & 65.84 (\textbf{+1.58}) \\
\hline
\end{tabular}
}
\label{Tab:offline_multimodal}
\end{table}
\end{document}